\documentclass{INTERSPEECH2023}

\interspeechcameraready

\title{Adapting an ASR Foundation Model for Spoken Language Assessment}
\name{Rao Ma, Mengjie Qian, Mark J. F. Gales, Kate M. Knill
\thanks{This paper reports on research supported by Cambridge University Press \& Assessment, a department of The Chancellor, Masters, and Scholars of the University of Cambridge. Mengjie Qian is supported by EPSRC Project EP/V006223/1 (Multimodal Video Search by Examples). Thanks to EST Ltd for initial analysis of Whisper on Linguaskill. }}
\address{ALTA Institute, Machine Intelligence Lab, Department of Engineering, Cambridge University, UK}
\email{\{rm2114,mq227,mjfg100,kmk1001\}@cam.ac.uk}

\usepackage{amsmath, amsfonts, tabu, subfigure, booktabs}
\usepackage{graphicx, bm}
\usepackage[toc,title,page]{appendix}
\usepackage{iitem}
\usepackage{multirow, amsfonts, amssymb}
\usepackage{appendix}
\usepackage[font=small,labelfont=bf]{caption}
\usepackage[dvipsnames]{xcolor}

\begin{document}

\maketitle
 
\begin{abstract}

A crucial part of an accurate and reliable spoken language assessment system is the underlying ASR model. Recently, large-scale pre-trained ASR foundation models such as Whisper have been made available. As the output of these models is designed to be human readable, punctuation is added, numbers are presented in Arabic numeric form and abbreviations are included. Additionally, these models have a tendency to skip disfluencies and hesitations in the output. Though useful for readability, these attributes are not helpful for assessing the ability of a candidate and providing feedback. Here a precise transcription of what a candidate said is needed. In this paper, we give a detailed analysis of Whisper outputs and propose two solutions: fine-tuning and soft prompt tuning. Experiments are conducted on both public speech corpora and an English learner dataset. Results show that we can effectively alter the decoding behaviour of Whisper to generate the exact words spoken in the response.

\end{abstract}
\noindent\textbf{Index Terms}: spoken language assessment, speech recognition, soft prompt tuning, ASR foundation model

\section{Introduction}
With the increasing worldwide population that is learning English as a second language (L2), there has been a growing interest in developing automatic spoken language assessment (SLA) systems for Computer Assisted Language Learning (CALL) applications~\cite{beatty2013teaching}. Compared to human evaluators, automatic assessment systems can provide consistent feedback to candidates in a timely and cost-effective manner. Automatic speech recognition (ASR) is usually the first component in a standard spoken language assessment system. 
The ASR model transcribes the candidate's response into readable text, which is then passed to other evaluation components to assess the candidate's proficiency in terms of vocabulary, fluency, topic development, etc.
Therefore, the quality of an automatic assessment system highly depends on the performance of the underlying ASR system. 

\begin{table}[htbp!]
    \centering
    \caption{Typical ASR error types made by Whisper: 1) {\color{red}abbreviation in red}; 2) {\color{blue} disfluency (false start and repetition) in blue}; 3) {\color{magenta}hesitation in pink}; 4); {\color{cyan}number in cyan} and 5) {\color{orange}partial word in orange}.}
    \begin{tabular}{l|l}
    \toprule
        Type & Sentence \\
        \midrule
        \multirow{3}*{Ref} & {\color{red} \textbf{mister}} lee when you arrive {\color{blue} \textbf{you could}} {\color{magenta}\textbf{uh}} we could \\
        & take {\color{blue}\textbf{the most}} the most cheap park zone blue zone \\
        & it costs {\color{magenta}\textbf{um}} {\color{cyan}\textbf{twenty dollar}} {\color{orange}\textbf{p-}} per week \\
        \midrule
        \multirow{3}*{Hyp} & {\color{red}\textbf{Mr.}} Lee, when you arrive, {\color{blue}*** *****} {\color{magenta}**} we could \\
        & take {\color{blue}*** ****} the most cheap Park zone, blue zone. \\
        & It costs {\color{magenta}**} {\color{cyan}\textbf{\$20} ******} {\color{orange}**} per week. \\
        \bottomrule
    \end{tabular}
    \label{tab:error_type}
    \vspace{-3mm}
\end{table}

Through the effort of researchers over decades, the ability of an ASR system has evolved from isolated word recognition \cite{cole1990speaker} into large vocabulary continuous speech recognition (LVCSR) \cite{graves2014towards, chan2016listen, gulati2020conformer}. 
Recently, large-scale ASR models pre-trained from large amounts of speech data have been made public. These foundation models~\cite{bommasanietal2022opportunities} demonstrate state-of-the-art performance on several speech recognition test sets \cite{baevski2020wav2vec, zhang2023google}. 
For instance, the model we chose for this paper Whisper~\cite{radford2022robust} learns from over 680K hours of labelled audio data collected from the Internet, enabling it to generate accurate hypotheses for both native and non-native speakers. Considering the success of high-performance foundation ASR models, it is therefore natural to integrate them into a spoken language assessment system. 

There are several defects, however, that we notice when applying Whisper to transcribe the candidate's speech. Rather than transcribing every word from the response into the corresponding spoken format, Whisper tends to generate punctuation, capitalisation, and phrases with inverse text normalisation (ITN) in decoding. Moreover, a large amount of deletions compared to the reference text have been observed on the test sets, which mostly correspond to repetitions and hesitations made by the candidate. This is due to the fact that Whisper is pre-trained on speech data collected from multiple sources, most of which have applied a high-level of ITN, e.g. numbers converted to Arabic numeral form, disfluencies removed from the speech, etc. Although this practice increases the readability of transcriptions for general purposes~\cite{radford2022robust}, it is unfavorable for a spoken language assessment system. For instance,  ``\texttt{uh}'',  ``\texttt{um}'', and the redundant ``\texttt{you could}'' in Table \ref{tab:error_type} are hesitation/extent indicators which should be kept for accurate assessment. 

One way to address this problem is to add an additional text processing step for the ASR output. 
We can easily remove the capitalisation and get rid of the unwanted punctuation in the generated hypotheses. For other token types such as numbers and abbreviations, a simple rule-based system \cite{akhtar2015iitp} or neural network based methods \cite{yolchuyeva2018text} can be adopted. These methods, however, add additional time costs and are prone to make errors in the prediction.
Also, the conversion to the original spoken words might be irreversible.
When ``\texttt{\$20}'' appears in the ASR output, we cannot possibly know whether the candidate said ``\texttt{twenty dollars}'' or ``\texttt{twenty dollar}'' in the first place, as mistakes in the plural form are common for beginners.
More importantly, it is impossible to restore the skipped hesitation and repetitions made by the candidate with text processing. 

In this paper, we propose two methods to ``undo'' the inherent post-processing within the Whisper model using small amounts of training data. The first method is model fine-tuning which is straightforward, yet computationally intensive. In addition, we propose a novel soft prompt tuning approach that is parameter-efficient. A few soft prompt embeddings are inserted in the decoder input and learned on the training data, leaving the original model parameters unchanged. 
Experiments show that with less than 20 hours data, we can largely reduce all types of formatting errors in the ASR hypotheses with both methods.

\section{Proposed Methods}

\subsection{Fine-tuning}
Denote the parameters of the ASR model as $\theta_{\text{ASR}}$. Suppose the acoustic features are $X = \{x_1, x_2, ..., x_n\}$, the ASR model is designed to find transcription $\hat Y = \{\hat y_1, \hat y_2, ..., \hat y_t\}$ such that
    $\hat Y = \mathop{\arg\max}_Y P(Y|X;\theta_{\text{ASR}})$.
Model fine-tuning (FT), where all model parameters are tuned with in-domain data to minimise the training loss, 
is a standard approach for domain adaptation. 
Here, we adopt the fine-tuning method to deal with the decoding issues of Whisper.
The Whisper model is pre-trained on a vast quantity of labelled audio-transcription data, which enables it to directly learn the speech recognition task in a supervised manner. Consequently, Whisper requires little additional fine-tuning to yield a well-performing ASR model for a specific domain. 

\subsection{Soft Prompts Tuning}


Within the Natural Language Processing (NLP) area, prompting \cite{liu2023pre, shin2020autoprompt} rather than fine-tuning \cite{mosbachstability} has become a popular method to adapt large language models (LLM) \cite{kenton2019bert, raffel2020exploring, OpenAI2023GPT4TR}. With prompting, we can provide initial context to model inputs to guide the generation process or design templates to transform model outputs. With this approach, the parameters of the pre-trained model remain unchanged. This makes prompting much more parameter-efficient than fine-tuning. In addition, we can prevent the model from over-fitting to the training data and forgetting general language knowledge. Prompting can be further classified into discrete prompts and soft prompts \cite{lester2021power, zhao2021discrete}.
The discrete prompt consists of natural language instructions that are carefully designed by a human \cite{schick2021s} or auto-generated using a LLM \cite{gao2021making}. For soft prompt tuning (SPT) \cite{lester2021power}, continuous vectors are inserted into the model input and optimised via gradient descent on the training data, which shows better model performance and requires minimal human expertise.

\begin{figure}[b] 
  \centering
  \vspace{-2mm}
  \subfigure{\includegraphics[width=1.0\linewidth]{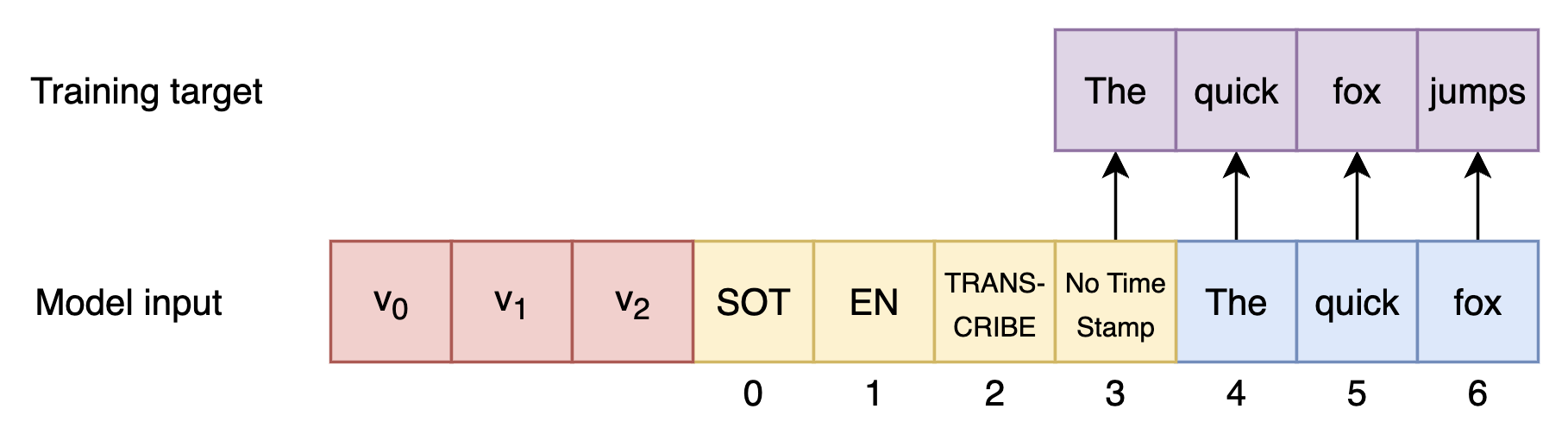}\label{fig:soft_prompt_training}}
  \vspace{-6mm}
  \caption{Modified decoder input with soft prompt training. Numbers refer to the order of positional embeddings.}
  \label{fig:soft_prompt_training_decoding}
\end{figure}

In the original method, soft prompt vectors are inserted into the encoder input of a LLM. However, for an ASR model, the decoder mostly captures the domain information from the reference text while the encoder learns audio-related features. Therefore, we propose to prepend $m$ token embeddings $V=\{v_0, ..., v_{m-1}\}$ to the decoder input embeddings. In training, only these parameters $V$ are updated based on the loss function while keeping the original model parameters, $\theta_\text{ASR}$, fixed. In this way, we add extra information for the model to condition on during the generation, 
$\hat Y = \mathop{\arg\max}_Y P(Y| X; \theta_{\text{ASR}}, \theta_V)$. Figure~\ref{fig:soft_prompt_training_decoding} illustrates the modified model input to the ASR decoder during training and decoding  when $m=3$.



\section{Evaluation Metrics}

\subsection{WER Calculation}
\label{sec:wer}
To evaluate the effectiveness of the proposed methods, we report the commonly used word error rate (WER) metric on the test sets. Different text normalisation methods are adopted before calculating WER. For \textbf{Raw WER}, we retrieve the raw decoding results from the ASR model and compare them to the reference text. Since the outputs of Whisper might vary significantly from the format of the reference text, the Raw WER can be fairly high. To better evaluate the performance of Whisper, \cite{radford2022robust} applies a text normalisation script\footnote{\url{https://github.com/openai/whisper/blob/main/whisper/normalizers/english.py}} to both the reference text and hypothesis before calculating WER, which we refer to here as \textbf{Standard WER}. The problem with this approach is that it changes the reference significantly and conceals the degree to which the hypothesis is similar to the original reference. Thus, we use a third \textbf{Speech WER} metric which mimics speech by removing punctuation and case from the reference and hypothesis. Since their presence but not form is needed, all hesitations are mapped to a \textit{\%hes\%} token. This makes it more suitable for evaluating the effectiveness of an ASR system for SLA. 


\begin{table}[htbp!]
    \centering
    \caption{Illustration of different WER metrics. Raw, Standard, and Speech WER results are 0.5, 0.0, and 0.33 respectively.}
    \label{tab:case}
    \begin{tabular}{l|l|l}
        \toprule
        Type & Metric & Sentence \\
        \midrule
        \multirow{3}*{Ref} & Raw & he bought um twenty ga- games \\
        & Standard & he bought 20 games \\
        & Speech & he bought \%hes\% twenty ga- games \\
        \midrule
        \multirow{3}*{Hyp} & Raw & He bought um 20 games. \\
        & Standard & he bought 20 games \\
        & Speech & he bought \%hes\% 20 games \\
        \bottomrule
    \end{tabular}
    \vspace{-3mm}
\end{table}

\subsection{Error Counts}
\label{sec:error_count}
Whisper possesses an inherent ITN process that inadvertently obscures the candidates' verbal mistakes. Through fine-tuning and soft prompt tuning, our aim is to ``undo'' the ITN and generate precise transcriptions of the spoken text. Within the language assessment task, various token types are of particular interest. 
To assess the effectiveness of the proposed methods in these aspects, we calculate the recall metric for these token types. This analysis provides insights into the improvement of the proposed approaches. To be specific, we analyze the following 5 types of words: 1) hesitation: words that represent a short pause in the sentence; 2) number: words that denote numbers; 3) abbreviation: words that most often appear in an abbreviation; 4) disfluency: comprises repetitions of words, false starts to a phrase, and discourse markers such as ``you know''~\cite{shriberg94}; 5) partial words: incomplete words uttered by the speaker. These word types are illustrated in Table~\ref{tab:error_type}.

For words belonging to each type, we count the occurrences $C_{all}$ in the reference. When the exact word as in the reference occurs in the ASR hypothesis, it is counted towards $C_{correct}$. It is worth noting that there can be an overlap among different word types. For example, ``forty nine forty nine'' encompasses both number and repetition types. Nevertheless, when we calculate the overall recall, i.e. $C_{correct} / C_{all}$, for all word types, each word is counted only once for both $C_{correct}$ and $C_{all}$.



\section{Experiments}

\subsection{Experimental Setup}
\label{sec:data}
We evaluate our proposed methods on three public datasets and one English language learner dataset, namely LibriSpeech~\cite{panayotov2015librispeech}, TED-LIUM 3~\cite{hernandez2018ted}, MGB3 dev17b~\cite{bell2015mgb} and Linguaskill~\cite{xu2020linguaskill}.
LibriSpeech is an audiobook-based English speech corpus, which covers a wide range of speakers with different accents.
TED-LIUM 3 is an audio dataset collected from TED talks, encompassing various topics such as science, education, and entertainment.
Multi-Genre Broadcast (MGB) comprises audio recordings from UK television and radio broadcasts, representing a variety of genres, including news, talk shows, and interviews. Linguaskill is recordings from an English language proficiency speaking test where candidates are second or foreign (L2) English learners\footnote{A public release of similar data is planned for 2023/4.}. A test submission includes various tasks including reading aloud, describing pictures, and speaking freely on a given topic. The general test focuses more on everyday conversation while the business test puts emphasis on business situations. There is no overlap of speakers in the split train, dev, and test sets. Table \ref{tab:statics} lists details of the test sets and our reproduced \textbf{Standard WER} results. Results on public datasets are comparable to \cite{radford2022robust}. 


\begin{table}[!htbp]
    \centering
    \caption{Statistics of test sets used in the experiments.}
    \begin{tabular}{lc|c|c|c}
    \toprule
        Dataset & Subset & \# Wds & Hours & WER \\
        \midrule
        \multirow{2}*{LibriSpeech} & test\_clean & 53K & 5.40 & 3.0 \\ 
        & test\_other & 52K & 5.34 & 6.7 \\ 
        TED-LIUM3 & test & 28K & 2.62 & 4.3 \\ 
        MGB3 & test & 52K & 4.63 & 13.3 \\ 
        \midrule
        \multirow{2}*{Linguaskill} & Ling\_general & 51K & 8.00 & 10.3 \\ 
        & Ling\_business & 47K & 8.00 & 13.9 \\
    \bottomrule
    \end{tabular}
    \label{tab:statics}
\end{table}


Various sizes of Whisper models have been released. The small.en model is used as the foundation model in these experiments due to its comparable performance to larger Whisper models and significantly faster runtime. Since we hope to examine the proposed methods when a small amount of labelled training data is available, a speech subset is randomly sampled from the training set of the target corpus. The Whisper model is tuned for 40 epochs or at most 30,000 steps on the selected training set. For both fine-tuning (FT) and soft-prompt tuning (SPT), 
a batch size of 20 is used for models trained on LibriSpeech and 5 for Linguaskill. 
For SPT, we run experiments with 1, 5, 20, or 100 soft prompts added. Models with 20 soft prompts perform the best in general, hence results are reported with $m=20$ unless otherwise stated. In the decoding, beam search with a width of 5 is adopted.

As described in Section~\ref{sec:wer}, the \textbf{Speech WER} metric is utilised in the following experiments. Further, we analyse ASR errors by calculating the recall for words belonging to specific types. 
We wrote rules to detect ``hesitation'', ``number'', and ``abbreviation'' types.
For Linguaskill, since ``disfluency'' (false starts, repetitions and discourse markers) and ``partial words'' have been tagged on the reference transcription by human annotators, we can calculate the counts for these two types. Such tags do not exist for standard datasets so we ignore the ``partial words'' type and find repeated words or phrases in the reference text with a regex command. Thus, ``repetitions" counted on standard datasets are not necessarily redundancies in speech.

\subsection{Results on Standard Datasets}

Table \ref{tab:lib_10h} shows the results on LibriSpeech. The baseline results are obtained by decoding the original Whisper model without further tuning. For both FT and SPT methods, 10 hours of data is used in training. 
On the in-domain test sets, FT and SPT yield average 15.5\% and 14.9\% improvement compared to the baseline system. The results indicate that the speech content can be more faithfully transcribed by the ASR system in the desired format. To study the ability of the approaches to generalise, we directly apply the SPT and FT models trained with 10h LibriSpeech to two additional standard test sets, TED-LIUM3 and MGB3, without any further training. Performance on TED-LIUM3 is significantly improved over the baseline. On MGB3, we observed that the performance of the model trained with SPT is  not stable for extremely short utterances. This might be due to the lack of such samples in the LibriSpeech training set.

\begin{table}[ht] 
    \centering
    \caption{Overall Speech WER on in-domain and out-of-domain test sets for models tuned on 10h LibriSpeech.}
    \label{tab:lib_10h}
    \begin{tabular}{@{ }l|cc|cc@{ }}
        \toprule
        \multirow{2}*{Model} & \multicolumn{2}{c|}{In-domain} & \multicolumn{2}{c}{Out-of-domain} \\
         & test\_clean & test\_other & TED-LIUM3 & MGB3 \\
        \midrule
        Baseline & 4.0 & 7.9 & 12.3 & 15.2 \\
        \midrule
        FT & 3.2 & 6.8 & 10.4 & 14.9 \\
        SPT & 3.2 & 6.9 & 9.8 & 15.4 \\
        \bottomrule
    \end{tabular}
\end{table}

Table~\ref{tab:error_count} presents the word occurrences for different token types in the reference $C_{all}$ and correctly predicted tokens in the transcribed text $C_{correct}$ for test\_other. Both FT and SPT show improvement on all four word types compared to the baseline, particularly in the number and abbreviation types. 
They each achieve a nearly 30\% absolute improvement in overall recall. Although the performance of SPT and FT are similar to each other on WER metrics and word recall, only 20 soft prompt embeddings are learned with SPT, contributing to a total of 15KB parameters. Yet the 15KB parameters achieve comparable performance to the FT counterpart that updates 244MB parameters.

\begin{table}[b] 
\vspace{-3mm}
    \centering
    \caption{Word counts and overall recall on test\_other.}
    \label{tab:error_count}
    \begin{tabular}{l|c|c|c|c}
    \toprule
     \multirow{2}*{Word Type} & $C_{all}$ & \multicolumn{3}{c}{$C_{correct}\uparrow$} \\
     & Ref & Baseline & FT & SPT \\
     \midrule
        Hesitation & 15 & 11 & 14 & 14 \\
        Number & 450 & 304 & 406 & 401 \\
        Abbreviation & 225 & 97 & 215 & 211 \\
        Repetition & 51 & 37 & 44 & 43 \\
    \midrule
    Recall All & - & 63.5\% & 92.5\% & 91.0\% \\
    \bottomrule
    \end{tabular}
\end{table}

Table~\ref{tab:error_cnt_all_tests} lists the overall recall for the four type of words on these different test sets. The improvement of recall is consistent for both in-domain and out-of-domain test sets.
The baseline recall for TED-LIUM3 and MGB3 are 45.3\% and 44.4\% respectively, which are lower than for LibriSpeech. SPT and FT models trained with 10h LibriSpeech data exhibit good generalisation capability on these out-of-domain test sets. A performance gain of over 30\% is observed on TED-LIUM3, with SPT achieving an absolute recall increase of 36\%. While the performance gain on MGB3 is not as substantial as that on TED-LIUM3, it still shows a promising improvement of 18\% absolute over the baseline. These results clearly indicate the models' capability to generalise effectively to unseen test sets.

\begin{table}[t] 
    \centering
    \caption{Overall word recall on in-domain and out-of-domain test sets for models tuned on 10h LibriSpeech.}
    \label{tab:error_cnt_all_tests}
    \begin{tabular}{@{ }l|cc|cc@{ }}
        \toprule
        \multirow{2}*{Model} & \multicolumn{2}{c|}{In-domain} & \multicolumn{2}{c}{Out-of-domain} \\
         & test\_clean & test\_other & TED-LIUM3 & MGB3 \\
        \midrule
        Baseline & 67.7\% & 63.5\% & 45.3\% & 44.4\% \\
        \midrule
        FT & 96.7\% & 92.5\% & 75.7\% & 62.1\% \\
        SPT & 96.0\% & 91.1\% & 81.3\% & 62.7\% \\
        \bottomrule
    \end{tabular}
\end{table}

\begin{figure}[htbp!]
    \centering
    \includegraphics[width=0.8\linewidth]{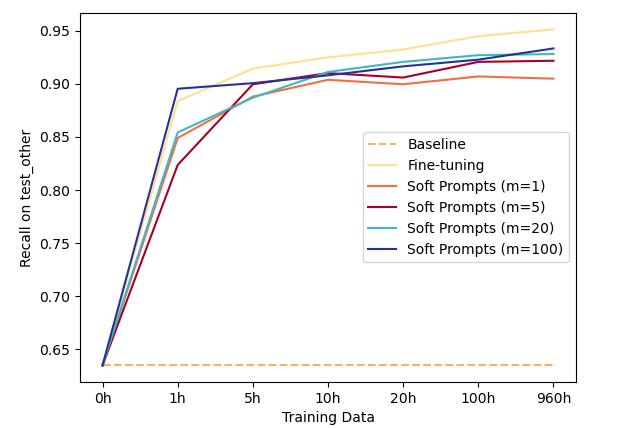}
    \caption{Overall word recall on test\_other when tuning on different hours of LibriSpeech training data.}
    \label{fig:lb_hour}
    \vspace{-3mm}
\end{figure}

As in-domain data is generally limited, we want to change the ASR behaviour with a small amount of training data. In  the above experiments, the models were trained with 10 hours of LibriSpeech data. To explore the impact of training data size on model performance, we trained models with varying quantities of data. Figure~\ref{fig:lb_hour} plots the overall word recall on the test\_other set with hours of training data. Even with a training set as small as 1 hour, both SPT and FT can achieve performance comparable to models trained with 960 hours. This demonstrates the effectiveness of our approaches in achieving substantial performance gains with a limited amount of training data.

\subsection{Results on L2 Learner Data}


\begin{table}[b] 
\setlength{\tabcolsep}{4.5pt}
    \centering
    \vspace{-2mm}
    \caption{Overall Speech WER results and breakdown of different error types for models tuned on 17h Linguaskill.}
    \label{tab:wer_lng}
    \begin{tabular}{@{ }l|ccc|c|ccc|c@{ }}
        \toprule
        \multirow{2}*{Model} & \multicolumn{4}{c|}{Ling\_general} & \multicolumn{4}{c}{Ling\_business} \\
        & Sub & Del & Ins & WER & Sub & Del & Ins & WER \\
        \midrule
        Baseline  & 4.5 & 10.7 & 1.3 & 16.4 & 5.8 & 16.6 & 1.4  & 23.9 \\
        \midrule
        FT  & 4.3 & 1.7 & 2.1 & 8.1 & 5.0 & 2.4 & 2.3 & 9.7 \\
        SPT & 4.4 & 1.7 & 2.8 & 8.9 & 5.3 & 2.5 & 3.2 & 11.0 \\
        \bottomrule
    \end{tabular}
    \vspace{-3mm}
\end{table}

Spoken responses collected from L2 learners generally encompass multiple hesitations, disfluencies, and partial words, which may differ from the token distribution in standard datasets  \cite{hilton2008link}. Moreover, these utterances are prone to grammatical errors or mispronunciations, increasing the difficulty of accurate recognition. Table~\ref{tab:wer_lng} presents the Speech WER results on Linguaskill test sets with breakdown of dirrerent error types. Here, 17 hours of Linguaskill training data is used in model tuning. FT and SPT significantly outperform the original baseline system by 55.7\% and 50.6\%. The table also shows that deletion errors are greatly reduced with FT and SPT. The results suggest that we can effectively recover the errors introduced in Whisper decoding.

As presented in Table~\ref{tab:error_cnt_lng}, the baseline Whisper model fails at generating words belonging to certain types that are of interest to language assessment, achieving an overall recall of only 15.4\%. Both SPT and FT significantly improve the model performance for all five word types, especially for hesitations, numbers, repetitions, and disfluencies. Since partial words contain words that are partly spoken and can have non-unique annotations, they are difficult for the model to learn and predict. In Table \ref{tab:case_analysis}, we show the decoding hypotheses of different models for one test utterance from Ling\_general. As can be seen, the ASR output becomes more accurate after FT or SPT.

\begin{table}[t] 
    \centering
    \caption{Word counts and overall recall on Ling\_general.}
    \label{tab:error_cnt_lng}
    \begin{tabular}{l|c|c|c|c}
    \toprule
     \multirow{2}*{Word Type} & $C_{all}$ & \multicolumn{3}{c}{$C_{correct}\uparrow$} \\
     & Ref & Baseline & FT & SPT \\
     \midrule
        Hesitation & 2661 & 5 & 2213 & 2267 \\
        Number & 421 & 220 & 388 & 381 \\
        Abbreviation & 18 & 17 & 17 & 17 \\
        Disfluency & 2201 & 583 & 1935 & 1938  \\
        Partial Words & 358 & 0 & 55 & 51 \\
    \midrule
    Recall All & - & 15.4\% & 82.1\% & 82.9\% \\
    \bottomrule
    \end{tabular}
    \vspace{-1mm}
\end{table}

\begin{table}[htbp!]
    \centering
    \caption{Case analysis on Ling\_general ({\color{red} \textbf{Errors in red}}).}
    \begin{tabular}{@{ }l@{ }|@{ }l@{ }}
    \toprule
        Type & Example \\
        \midrule
        \multirow{2}*{Ref} & \footnotesize{\%hes\% i think i'm not i'm not really denominal maybe} \\
        & \footnotesize{\%hes\% one hundred because i'm not i'm not like shopping}\\
        \midrule
        \multirow{2}*{Baseline} & \footnotesize{{\color{red}*****} i think i'm not i'm not really {\color{red}\textbf{the nominal}} maybe }\\
        & \footnotesize{{\color{red}*****} {\color{red}\textbf{a 100}} because {\color{red}*** ***} i'm not like shopping} \\
        \midrule
        \multirow{2}*{SPT} & \footnotesize{\%hes\% i think i'm not i'm not really {\color{red}\textbf{the nominal}} maybe} \\
        &\footnotesize{\%hes\% one hundred because i'm not i'm not like shopping} \\
        \midrule
        \multirow{2}*{FT} & \footnotesize{\%hes\% i think i'm not i'm not really denominal maybe} \\
        & \footnotesize{\%hes\% one hundred because i'm not i'm not like shopping} \\
        \bottomrule
    \end{tabular}
    \label{tab:case_analysis}
    \vspace{-3mm}
\end{table}

\section{Conclusions}
Large-scale ASR foundation models are becoming increasingly popular and available. 
In this paper, we discuss the problem of applying such a model (here Whisper) to a spoken language assessment system. 
Whisper has a tendency to omit disfluencies and hesitations in its output, which may be harmful to accurate candidate assessment.
To address this, we propose model fine-tuning and a novel soft prompt tuning approach. In soft prompting, the original model parameters are fixed  and a small number of additional prompt parameters are estimated. This approach is efficient as the model  is not modified, and yields comparable performance to  fine-tuning. Applying either soft-prompting or fine-tuning enables Whisper to generate decoding output that should be better tuned for use in spoken language assessment.


\bibliographystyle{IEEEtran}
\bibliography{mybib}

\end{document}